\documentclass[10pt,twocolumn,letterpaper]{article}

\usepackage{iccv}
\usepackage{times}
\usepackage{epsfig}
\usepackage{graphicx}
\usepackage{tabularx}
\usepackage{amsmath,mathtools}
\usepackage{amssymb}
\usepackage[inline]{enumitem}
\usepackage[dvipsnames]{xcolor}

\usepackage{booktabs}
\usepackage{tabularx}
\usepackage{multirow}
\usepackage{colortbl}
\usepackage{subfig}
\usepackage{float}

\usepackage{algpseudocode,algorithm,algorithmicx}
\usepackage{xspace}
\usepackage{interval}
\usepackage{bm,upgreek}
\usepackage[pagebackref=true,breaklinks=true,letterpaper=true,colorlinks,citecolor=ForestGreen, bookmarks=false]{hyperref}

\def\HiLi{\leavevmode\rlap{\hbox to 0.85\linewidth{\color{gray!20}\leaders\hrule height .8\baselineskip depth .5ex\hfill}}}

\makeatletter
\newcommand{\algcolor}[2]{%
  \hskip-\ALG@thistlm\colorbox{#1}{\parbox{\dimexpr\linewidth-2\fboxsep}{\hskip\ALG@thistlm\relax #2}}%
}

\makeatother

\makeatletter
\newcommand*\wt[1]{\mathpalette\wthelper{#1}}
\newcommand*\wthelper[2]{%
        \hbox{\dimen@\accentfontxheight#1%
                \accentfontxheight#11.3\dimen@
                $\m@th#1\widetilde{#2}$%
                \accentfontxheight#1\dimen@
        }%
}

\newcommand*\accentfontxheight[1]{%
        \fontdimen5\ifx#1\displaystyle
                \textfont
        \else\ifx#1\textstyle
                \textfont
        \else\ifx#1\scriptstyle
                \scriptfont
        \else
                \scriptscriptfont
        \fi\fi\fi3
}
\makeatother

\newcommand{\synthia}{{\scshape Synthia}\xspace}

\newcommand{\system}{ACE\xspace}

\newcommand{\ra}[1]{\renewcommand{\arraystretch}{#1}}
\newcommand{\highway}{{\scshape Highway}\xspace}
\newcommand{\nyc}{{\scshape NYC-like City}\xspace}
\newcommand*\rot{\rotatebox{90}}
\definecolor{Gray}{gray}{0.85}
\newcolumntype{g}{>{\columncolor{Gray}} c}

\newcommand{\zz}{{\bm z}}

\newcommand{\kl}[2]{{\texttt{KL}}\left(#1\;\middle\|\;#2\right)}

\newcommand{\task}{\mathcal{T}}

\newcommand{\data}{\mathcal{D}}
\newcommand{\inp}{\bm{x}^s}
\newcommand{\out}{\bm{y}^s}

\newcommand{\buffer}{\mathcal{M}}

\newcommand{\cL}{\mathcal{L}}

\newcommand{\param}{{\bm{\uppsi}}}               %
\newcommand{\udparam}{\wt{\param}}     %

\usepackage{xcolor}

\graphicspath{{./figures/}}

\iccvfinalcopy %

\def\iccvPaperID{XXX} %

\ificcvfinal\fi
\begin{document}

\title{ACE: Adapting to Changing Environments for Semantic Segmentation}

\author{Zuxuan Wu$^1$, Xin Wang$^2$, Joseph E. Gonzalez$^2$, Tom Goldstein$^1$, Larry S. Davis$^1$ \\ \\
$^1$University of Maryland, ~~$^2$UC Berkeley
}

\maketitle

\begin{abstract}
Deep neural networks exhibit exceptional accuracy when they are trained and tested on the same data distributions. However, neural classifiers are often extremely brittle when confronted with domain shift---changes in the input distribution that occur over time. We present ACE, a framework for semantic segmentation that dynamically adapts to changing environments over the time. By aligning the distribution of labeled training data from the original source domain with the distribution of incoming data in a shifted domain, ACE synthesizes labeled training data for environments as it sees them.  This stylized data is then used to update a segmentation model so that it performs well in new environments. To avoid forgetting knowledge from past environments, we introduce a memory that stores feature statistics from previously seen domains.  These statistics can be used to replay images in any of the previously observed domains, thus preventing catastrophic forgetting. In addition to standard batch training using stochastic gradient decent (SGD), we also experiment with fast adaptation methods based on adaptive meta-learning. Extensive experiments are conducted on two datasets from SYNTHIA, the results demonstrate the effectiveness of the proposed approach when adapting to a number of tasks. 
\end{abstract}

\section{Introduction}
When computer vision systems are deployed in the real world, they are exposed to changing environments and non-stationary input distributions that pose major challenges.  For example, a deep network optimized using images collected on sunny days with clear skies may fail drastically at night under different lighting conditions. In fact, it has been recently observed that deep networks demonstrate severe instability even under small changes to the input distribution~\cite{hendrycks2018benchmarking}, let alone when confronted with dynamically changing streams of information. 
\begin{figure}[t!]
\begin{center}
   \includegraphics[width=0.82\linewidth]{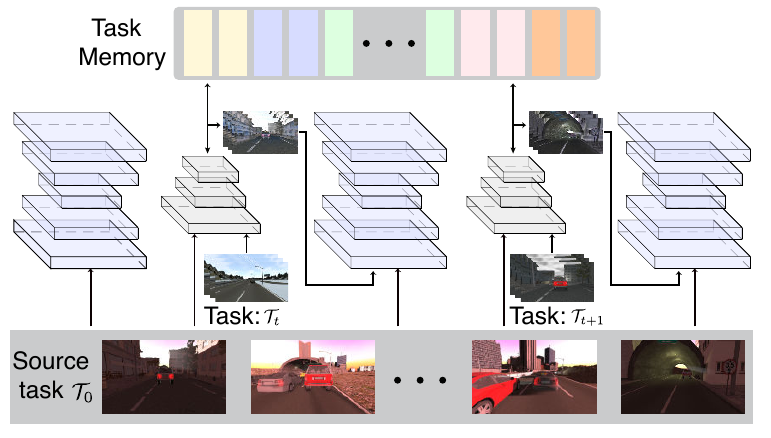}
\end{center}
\vspace{-0.2in}
   \caption{\textbf{A conceptual overview of the framework}. \system adapts a model trained on a source task to a sequence of target tasks.  This is done by aligning the feature statistics of labeled images from a source task with incoming images of the target task.  This alignment produces labeled images in the target domain that can be used to update the segmentation model. A memory unit also facilitates replay of images from domains seen in the past to prevent forgetting.}
\label{fig:framework}
\end{figure}

The problem of domain shift can be avoided by collecting sufficient training data to cover all possible input distributions that occur at test time. However, the expense of collecting and manually annotating data makes this infeasible in many applications.  This is particularly true for detailed visual understanding tasks like object detection and semantic segmentation where image annotation is labor-intensive. It is worth noting that humans are capable of ``lifelong learning,'' in which new tasks and environments are analyzed using accumulated knowledge from the past. However, achieving the same goal in deep neural networks is non-trivial as (i) new data domains come in at real time without labels, and (ii) deep networks suffer from \emph{catastrophic forgetting}~\cite{mccloskey1989catastrophic}, in which performance drops on previously learned tasks when optimizing for new tasks.

We consider the lifelong learning problem of adapting a pre-trained model to dynamically changing environments, whose distributions reflect disparate lighting and weather conditions. In particular, we assume access to image-label pairs from an original {\em source} environment, and only unlabeled images from new {\em target} environments that are not observed in the training data.  Furthermore, we consider the difficulties posed by learning over time, in which target environments appear sequentially.  

We focus on the specific task of semantic segmentation due to its practical applications in autonomous driving, where a visual recognition system is expected to deal with changing weather and illumination conditions. This application enables us to leverage the convenience of collecting data from different distributions using graphic rendering tools~\cite{ros2016cvpr,richter2016playing}.

To this end, we introduce \system, a framework which adapts a pre-trained segmentation model to a stream of new tasks that arrive in a sequential manner, while storing historical style information in a compact memory to avoid forgetting. 
In particular, given a new task, we use an image generator to align the distribution of (labeled) source data with the distribution of (unlabeled) incoming target data at the pixel-level.  This produces labeled images with color and texture properties that closely reflect the target domain, which are then directly used for training the segmentation network on the new target domain. Style transfer is achieved by renormalizing feature maps of source images so they have first- and second-order feature statistics that match target images~\cite{huang2017adain,DBLP:journals/corr/UlyanovVL16}.  These renormalized feature maps are then fed into a generator network that produces stylized images.
 
What makes \system unique is its ability to learn over a lifetime.  To prevent forgetting, \system contains a compact and light-weight memory that stores the feature statistics of different styles.  These statistics are sufficient to regenerate images in any of the historical styles without the burden of storing a library of historical images.  Using the memory, historical images can be re-generated and used for training throughout time, thus stopping the deleterious effects of catastrophic forgetting. The entire generation and segmentation framework can be trained in a joint end-to-end manner with SGD. Finally, we consider the topic of using adaptive meta-learning to facilitate faster adaptation to new environments when they are encountered.

Our main contributions are summarized as follows: (1) we present a lightweight framework for semantic segmentation, which is able to adapt to a stream of incoming distributions using simple and fast optimization; (2) we introduce a memory that stores feature statistics for efficient style replay, which facilitates generalization on new tasks without forgetting knowledge from previous tasks; (3) we consider meta-learning strategies to speed up the rate of adaptation to new problem domains; (4) we conduct extensive experiments on two subsets of \synthia~\cite{DBLP:journals/corr/RosSAW16} and the experiments demonstrate the effectiveness of our method when adapting to a sequence of tasks with different weather and lighting conditions.
\section{Related Work}
\noindent\textbf{Unsupervised Domain Adaptation}. Our work relates to unsupervised domain adaptation, which aims to improve the generalization of a pre-trained model when testing on novel distributions without accessing labels. Existing approaches along this line of research to reduce domain differences at either the feature or pixel level. In particular, feature-level adaptation focuses on aligning feature representations used for the target task (\eg, classification or segmentation) by minimizing a notion of distance between source and target domains. Such notion of distance can be explicit metrics in the forms of Maximum Mean Discrepancies (MMD)~\cite{DBLP:conf/icml/LongC0J15,bousmalis2016domain}, covariances~\cite{DBLP:conf/eccv/SunS16}, \etc; or implicitly estimated to make features domain-invariant using adversarial loss functions such as reversed gradient~\cite{DBLP:conf/icml/GaninL15,ganin2016domain}, domain confusion~\cite{tzeng2015simultaneous}, or Generative Adversarial Network ~\cite{DBLP:conf/cvpr/TzengHSD17,hoffman2016fcns,hong2018conditional,saito2018maximum,huang2018domain}, \etc. 

\begin{figure*}[t!]
\begin{center}
   \includegraphics[width=0.95\linewidth]{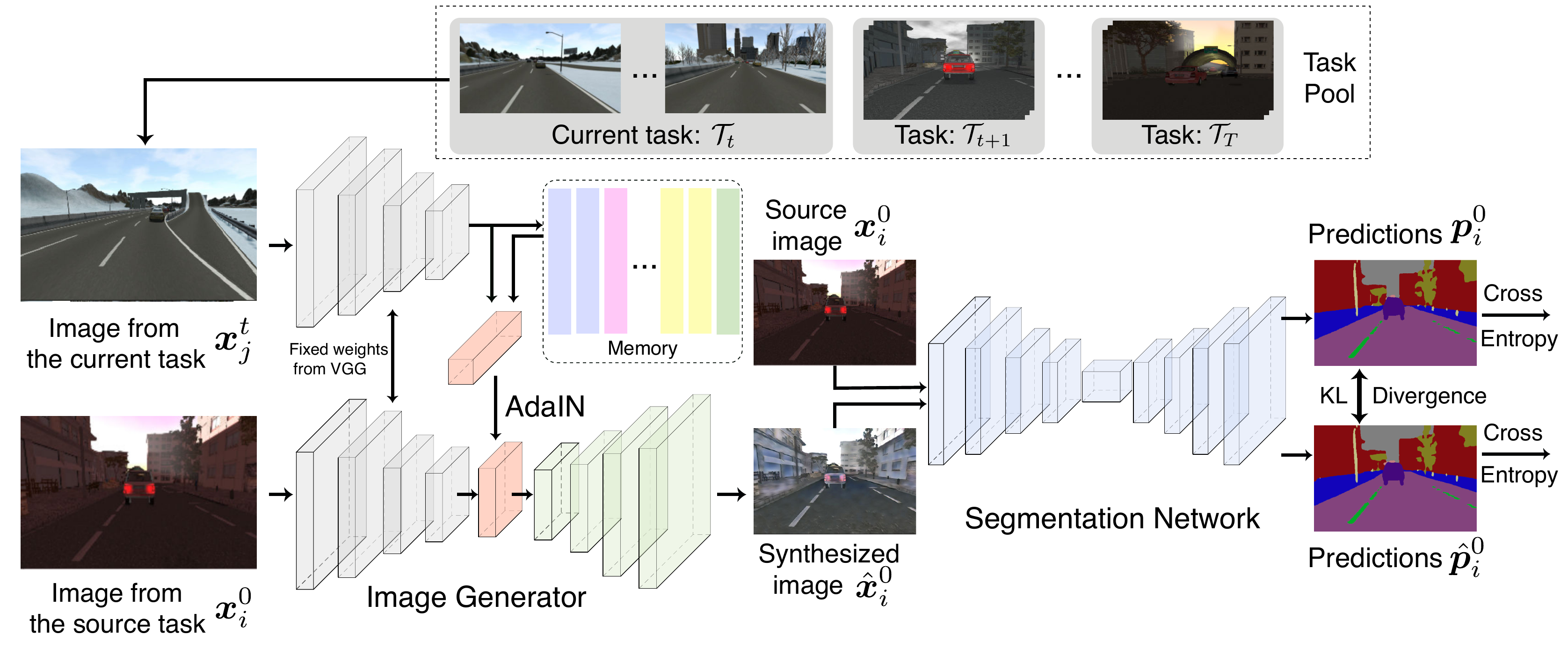}
\end{center}
\vspace{-0.2in}
   \caption{\textbf{An overview of the proposed framework}. Given an incoming task, \system synthesizes new images that preserve the contents of images from the source task but in the style of a target task.  This is done either by transferring style information from incoming images onto the source images, or by sampling style information from the memory unit. With these synthesized images in different styles, the segmentation network is trained to generalize to new tasks without forgetting knowledge learned in the past.}
\label{fig:framework}
\end{figure*}

On the other hand, pixel-level adaptation transforms images from different domains to look as if they were drawn from the same distribution by using a mapping that reduces inconsistencies in texture and lighting~\cite{DBLP:conf/cvpr/BousmalisSDEK17,DBLP:conf/cvpr/ShrivastavaPTSW17,DBLP:journals/corr/TaigmanPW16,DBLP:conf/nips/LiuT16}. There are also recent methods seeking to align both pixel-level and feature-level representations simultaneously~\cite{hoffman2017cycada,wu2018dcan,zhang2018fully}. In addition, Zhang \etal introduce a curriculum strategy that uses global label distributions and local super-pixel distributions for adaptation. Saleh \etal handle foreground classes using detection methods when addressing domain shift~\cite{saleh2018effective}. Our framework differs from previous work as we are adapting to a stream of testing domains that arrive sequentially rather than a single fixed one, which is challenging as it requires the network to perform well on both current and all previous domains. Note that although we mainly focus on pixel-level alignment, our method can further benefit from feature-level alignment in the segmentation network, but at the cost of saving raw images as opposed to only feature statistics. Further, our approach is also related to~\cite{wulfmeier2018incremental,bobu2018adapting,hoffman2014continuous} that perform sequential adaptation for classification tasks by aligning at feature-level, while ours focuses on semantic segmentation with alignment at pixel-level.

\vspace{0.03in}
\noindent\textbf{Image Synthesis and Stylization}. There is a growing interest in synthesizing images with Generative Adversarial Networks (GANs)~\cite{DBLP:conf/eccv/YooKPPK16,radford2015unsupervised,DBLP:conf/nips/LiuT16}, which is formulated as a minimax game between a generator and a discriminator~\cite{goodfellow2014generative}. To control the generation process, a multitude of additional information has been incorporated including labels~\cite{odena2016conditional}, text~\cite{reed2016generative}, attributes~\cite{shen2016learning}, and images~\cite{DBLP:conf/cvpr/IsolaZZE17,ledig2016photo}. GANs have also been used in the context of image-to-image translation, which transfers the style of an image to that of a reference image using either cycle-consistency~\cite{DBLP:conf/iccv/ZhuPIE17} or mapping into a shared feature space~\cite{liu2017unsupervised,huang2018multimodal}. Without knowing joint distributions of domains, these approaches attempt to learn conditional distributions with marginal distributions from each domain. However, generating high resolution images with GANs still remains a difficult problem and is computationally intensive~\cite{karras2017progressive}. In contrast, methods for neural style transfer~\cite{gatys2016image,huang2017adain,ulyanov2016texture,perarnau2016invertible,johnson2016perceptual} usually avoid the difficulties of generative modeling, and simply match the feature statistics of Gram matrices~\cite{gatys2016image,johnson2016perceptual} or perform channel-wise alignment of mean and variance~\cite{li2017demystifying,huang2017adain}. In our work, we build upon style transfer to synthesize new images in the style of images from the current task while preserving the contents of the source image. 

\vspace{0.03in}
\noindent\textbf{Lifelong Learning}. Our work is also related to lifelong learning, or continual learning, which learns progressively and adapts to new tasks using knowledge accumulated throughout the past. Most existing work focuses on mitigating catastrophic forgetting when learning new tasks~\cite{kirkpatrick2017overcoming,zenke2017continual,rebuffi2017icarl,shin2017continual,shmelkov2017incremental,lopez2017gradient,Castro_2018_ECCV}. Several recent approaches propose to dynamically increase model capacities when new tasks arrive~\cite{yoon2017lifelong,xu2018reinforced}. Our work focuses on how to adapt a learned segmentation model in an unsupervised manner to a stream of new tasks, each with image distributions different from those originally used for training. In addition, to avoid forgetting knowledge learned in the past, styles are represented and cataloged using their feature statistics.  Because this representation is much smaller than raw images, the framework is scalable.

\vspace{0.03in}
\noindent\textbf{Meta-Learning}. Meta-learning~\cite{schmidhuber1997shifting,thrun1998learning}, also known as learning to learn, is a setting where an agent ingests a set of tasks, each a learning problem on its own, and then establishes a model that can be quickly adapted to unseen tasks from the same distribution. There are three categories of meta-learners: (i) model-based with an external memory~\cite{santoro2016meta,munkhdalai2017meta}; (ii) metric-based~\cite{vinyals2016matching}; and (iii) optimization-based~\cite{finn2017model,reptile}. Existing approaches mainly focus on few shot classification, regression, and reinforcement learning problems, while our approach focuses on how to adapt segmentation models efficiently. 

\section{Approach}
The goal of \system is to adapt a segmentation model from a source task to a number of sequentially presented target tasks with disparate image distributions. The method transfers labeled source images into target domains to create synthetic training data for the segmentation model, while memorizing style information to be used for style replay to prevent forgetting.

More formally, let $\task_0$ denote the source task and $\{\task_i\}^T_{i=0}$ represent $T$ target tasks that arrive sequentially. We further use ${\bm X}^0 = \{({\bm x}^0_1, {\bm y}^0_1), \cdots, ({\bm x}^0_N, {\bm y}^0_N)\}$~\footnote{We omit $\task$ here for the ease of notation.} to represent the $N$ images and their corresponding labels used for the source task. The label ${\bm y}^0_i$ contains a ones-hot label vector for each pixel in the image ${\bm x}^0_i$;
 we denote the $i$-th image sample as  ${\bm x}^0_i \in \mathbb{R}^{3 \times H \times W },$  and ${\bm y}^0_i \in\{0,1\}^{C \times H \times W}$ represents the corresponding label maps, with $H$ and $W$ being the height and width respectively and $C$ denoting the number of classes.

For each subsequent target task, we assume access to only images rather than image-label pairs as in the source task. We further denote the number of target tasks as $T$ and use ${\bm X}^t = \{{\bm x}^t_1, \cdots, {\bm x}^t_{N^t}\}$ for $t \in [1 \cdots T]$ to represent the image set for the $t$-th incoming task, which has $N^t$ images of the same resolution as the source data.

\system contains four key components: an encoder, a generator, a memory, and a segmentation network.  The encoder network converts a source image $\bm x^0_i$ into a feature representation $\bm z^0_i$ (in our case, a stack of 512 output feature maps).  The generator network converts feature representations $\bm z$ into images.  The style of the resulting image can be controlled/manipulated by modifying the statistics (mean and standard deviation of each feature map) of $\bm z$ before it is handed to the generator.  The memory unit remembers the feature statistics (1024 scalar values per style, corresponding to the mean and standard deviation of each of the 512 feature maps) for each image style/domain.  A source image can be stylized into any previously seen domain by retrieving the relevant style statistics from the memory unit, renormalizing the feature maps of the source image to have the corresponding statistics, and then handing the renormalized features to the generator to create an image.

\vspace{0.05in}
\noindent\textbf{Stylization via the encoder and generator}. When a new task is presented, labeled images are created in the new task domain by transferring  source images (and their accompanying labels) to the target domain.  To do this, we jointly train a generator network for producing target-stylized images, and a segmentation network for processing images in the target domain.

The image generation pipeline begins with an encoder that extracts feature maps from images.   We use a pre-trained VGG19 network~\cite{simonyan2014very} as the encoder, taking the output from \texttt{relu4} to define $f_{enc}$. Following~\cite{li2017universal,huang2017adain}, the weights of the encoder are frozen during training to extract fixed representations $f_{enc}( {\bm x}^0_i)=\zz_i^0$ and $f_{enc}({\bm x}^t_j) =\zz_j^t$ from images ${\bm x}^0_i$ and ${\bm x}^t_j$, respectively.   

The image generator $f_{gen}$, parameterized by weights $\param_{gen}$, de-convolves feature maps into images. The style of the output image can be borrowed from a target image with \textrm{AdaIN}~\cite{huang2017adain}, which re-normalizes the feature maps (\ie, channels) $\zz_i^0$ of source images to have the same mean and standard deviation as the maps of a selected target image $\zz_i^t$:
\begin{align}
\label{eqn:adain}
	\hat{\zz}_i^0 = \textrm{AdaIN}(\zz_{i}^0,\zz_{j}^t) = \sigma(\zz^{t}_{j}) \frac{\zz_{i}^0-\mu(\zz_{i}^0)}{\sigma(\zz_{i}^0)} + \mu(\zz^t_{j}).
\end{align}
Here, $\sigma(\zz)$ and $\mu(\zz)$ compute the mean and variance of each channel of $\zz$, respectively. 
The normalized feature maps $\hat{\zz}_i^0$ can be stuffed into the generator to synthesize a new image $\hat{\bm x}_i^0 = f_{gen}(\hat{\zz}_i^0; \param_{gen})$. If the parameters $\param_{gen}$ are properly tuned, the resulting image will have the contents of ${\bm x}^0_i$ but in the style of ${\bm x}^t_j$. 

We train the generator so that it acts as an inverse for the encoder;  the encoder should map the decoded image (approximately) onto the features that produced it.  We enforce this by minimize the following loss function:
\begin{equation}
\begin{aligned}
\ell_{gen}(\param_{gen}) =  & \lVert   \tilde \zz   - \hat{\zz}_i^0 \rVert_{2} +  
  \lVert  \mu( \tilde{\zz}) - \mu(\zz_j^t)  \rVert_{2} \\
 &  \qquad + \lVert   \sigma( \tilde{\zz})-\sigma(\zz_j^t)  \rVert_{2},\\
 \text{ where }  &  \tilde{\zz} = f_{enc}(  f_{dec}(  \hat{\zz}_i^0 ; \param_{gen} ) ). 
\label{eq:sty}
\end{aligned}
\end{equation}
Here, the first term (the content loss) measures the differences between features of the generated image and the aligned features of the source image with an aim to preserve the contents the source images. The remaining two terms force the generated image into the style of ${\bm x}^t_j$ by matching the mean and variance of feature maps per-channel. Note that some authors match gram matrices~\cite{gatys2016image,wu2018dcan} to make styles consistent. We match mean and variances of feature maps as in~\cite{li2017demystifying,ulyanov2016texture} since these statistics are simple to optimize and contain enough information to get a good stylization. In contrast to using several layers for alignment~\cite{li2017demystifying,huang2017adain}, we simply match one layer of feature maps from the VGG encoder, which is faster yet sufficient. More importantly, this facilitates lightweight style replay as will be described below.

\noindent\textbf{The segmentation network}.
The newly synthesized image $\hat{\bm x}_i^0 = f_{dec}(   \tilde{\zz}  ; \param_{gen} ) $ is handed to the segmentation network $f_{seg}$, parameterized by weights $\param_{seg}$. This network produces a map of label vectors $\hat{\bm p}_i^0 = f_{seg}(\hat{\bm x}_i^0;\param_{seg}),$ and is trained by minimizing a multi-class cross-entropy loss summed over pixels. In addition, since the synthesized image might lose certain details of the original images that could degrade the performance of the segmentation network, we further constrain outputs of the synthetic image $\hat{\bm x}_i^0$ from the segmentation network $\hat{\bm p}_i^0$ to be close to the predictions ${\bm p}_i^0$ of the original image ${\bm x}_i^0$ before stylization. This is achieved by measuring the $\texttt{KL}$-divergence between these two outputs, which is similar in spirit to knowledge distillation~\cite{DBLP:journals/corr/HintonVD15} with the outputs from the original image serving as the teacher. The segmentation loss takes the following form:
\begin{align}
\begin{split}
\ell_{seg}(\param_{seg}) =& - \sum^{H\times W}_{m=1}  \kl{\hat{\bm p}_{i,m}^0}{{\bm p}_{i,m}^0}  \\
  &+ \sum_{c=1}^{C} {\bm y}^0_{i,mc}\texttt{log}(\hat{\bm p}^0_{i,mc}).
\label{eq:seg}
\end{split}
\end{align}

Finally, combining Eqn.~\ref{eq:sty} and Eqn.~\ref{eq:seg}, we have the following objective function:
\begin{align}
\hspace{-2mm}\mathcal{L}(\param) =  \underset{\substack{(\bm{x}^0, \, \bm{y}^0) \thicksim {\bm X}^0 \\ {\bm x}^t \thicksim {\bm X}^t }}{ \mathbb{E}}   
   \, \ell_{gen}({\bm x}^0, {\bm x}^t) + \ell_{seg}({\bm x}^0, {\bm y}^0, {\bm x}^t),
\label{eq:obj}
\end{align}
where $\param = \{\param_{seg}, \param_{gen}\}$ denotes the parameters of the network.  Note that the segmentation loss implicitly depends on the generator parameters because segmentation is performed on the output of the generator.

\vspace{0.05in}
\noindent \textbf{Memory unit and style replay}. Optimizing Eqn~\ref{eq:obj} reduces the discrepancies between the source task and the target task, yet it is unclear how to continually adapt the model to a sequence of incoming tasks containing potentially different image distributions without forgetting knowledge learned in the past. A simple way is to store a library of historical images from previous tasks, and then randomly sample images from the library for replay when learning new tasks. However, this requires large working memory which might not be viable, particularly for segmentation tasks, where images are usually of high resolutions (\eg, $1024 \times 2048$ for images in Cityscapes~\cite{richter2016playing}). 

Fortunately, the proposed alignment process in Eqn.~\ref{eqn:adain} synthesizes images from the target distribution using only a source image, and the mean and variance of each channel in the feature maps $\zz_j^t$ from a target image. Therefore, we only need to save the feature statistics ($512$-D for both mean and variance) in the memory $\mathcal{M}$ for efficient replay. When learning the $t$-th task $\task_t$, we select a sample of test images and store their $1024$-D feature statistics in the memory.  When adapting to the next task $\task_{t+1}$, in addition to sampling from ${\bm X}^t$, we also randomly access the memory $\mathcal{M},$ which contains style information from previous tasks, to synthesize images that resemble seen tasks on-the-fly for replay. 

\label{sec:reptile}
\vspace{0.05in}
\noindent \textbf{Faster adaptation via adaptive meta-learning}.
 Recent methods in meta-learning~\cite{finn2017model,reptile,ravi2016optimization} produce flexible models having meta-parameters with the property that they can be quickly adapted to a new task using just a few SGD updates.   While standard SGD offers good performance when optimizing Eqn.~\ref{eq:obj}  for a sufficient number of epochs, we now explore whether adaptive meta-learning can produce models that speed up adaptation. 

For this purpose, we use Reptile~\cite{reptile}, which is an inexpensive approximation of the MAML~\cite{finn2017model} method.  Reptile updates parameters of a meta-model by first selecting a task at random, and performing multiple steps of SGD to fine-tune the model for that task. Then a ``meta-gradient'' step is taken in the direction of the fine-tuned parameters.  The next iteration proceeds with a different task, and so on, to generate a meta-model with parameters that are only a small perturbation away from the optimal parameters for a wide range of tasks. 

To be precise, the Reptile meta gradient $\mathbf{g}_t(\param)$ is defined as:
\begin{equation} 
\begin{aligned}
\label{eq:grad}
    \mathbf{g}_t(\param) & = \param_t - \udparam, \ \text{where} \, \ \udparam = U^k(\param_t).
\end{aligned}
\end{equation}
Here $U^k(\param_t)$ denotes $k$ steps of standard SGD for a randomly selected task. To achieve fast adaptation, we sample from the current task as well as the memory to perform meta-updates using meta-gradients from the whole history of tasks. The meta-gradients are then fine-tuned on the current task to evaluate performance. The algorithm is summarized in Alg.~\ref{alg:metaace}.

\begin{algorithm}[t]
\caption{Fast Adaptation with Adaptive Meta-Learning}
\label{alg:metaace}
\begin{algorithmic}[1]
{\footnotesize
\State {\bfseries Input:} ${\bm X}^0 = \{({\bm x}^0_1, {\bm y}^0_1), \cdots, ({\bm x}^0_N, {\bm y}^0_N) \}$ 
\State A pre-trained segmentation model, whose parameters are $\param$
\State The memory is inialized as empty $\buffer \leftarrow [~]$
\For{$t = 1,\dots T$}
\State initialize $\data_{t} = \emptyset$
\While{$| \data_{t}| < N^t$}
\For{$\texttt{iterations} =1, \dots, I$}
\State Append batch of $n$ image samples ${\bm x}^t$ to $\data_{t}$
\State Sample batches of $\left( \inp, \out \right)$ from the source task  
\State Sample batches of $\bm{x}^t$ from the $t$-th task  
\State Compute $\bm{z}^t$ with the VGG encoder using sampled $\bm{x}^t$
\State Sample seen tasks $\bm{z}^l (l<t)$ from the memory $\buffer$
\State \HiLi{$\param_{t} \leftarrow \param_t - \eta\,\mathbf{g}_t(\param)$, \, $\mathbf{g}_t(\param)$ are computed with Eqn.~\ref{eq:grad}}
\State Update $\buffer$ by storing some randomly selected $\bm{z}^t$
\EndFor
\State  {$\udparam_{t} \leftarrow \param_t - \alpha \nabla \cL(\param_t;\data_t)$} \Comment{For testing the $t$-th task }
\EndWhile
\State  {$\param_{t+1} \leftarrow \param_t$}
\EndFor
}
\end{algorithmic}
\end{algorithm}

\section{Experiments}
In this section, we first introduce the experimental setup and implementation details. Then, we report results of our proposed framework on two datasets and provide some discussions.

\begin{table*}[t!]
\setlength{\tabcolsep}{2pt}
\resizebox{\linewidth}{!}{
\ra{1.2}
\begin{tabular}{@{}*{13}cgg*{11}cgg@{}}
\toprule
&& && \multicolumn{10}{c@{}}{\highway} & \multicolumn{14}{c@{}}{\nyc} \\

\cmidrule{5-15} \cmidrule{17-28}
Method && \rot{Arch.} && \rot{Dawn} & \rot{Fall} &  \rot{Fog} &  \rot{Night} &  \rot{Spring} &  \rot{Summer} &  \rot{Sunset} &  \rot{Winter} &  
\rot{WinterNight} &   \rot{mean mIOU} &  \rot{gain} && \rot{Dawn} & \rot{Fall} &  \rot{Fog} &  \rot{Night} &  \rot{RainNight} & \rot{SoftRain} & \rot{Spring} &  \rot{Summer} &  \rot{Sunset} &  \rot{Winter} &   \rot{mean mIOU}  & \rot{gain} \\

\cmidrule{1-1} \cmidrule{3-3} \cmidrule{5-15} \cmidrule{17-28}

Source  && A && 65.4 & 61.4 & 62.3 & 59.4 & 62.3 & 62.1 & 64.9 & 50.0 & 54.5 & 60.2 &- & 
& 46.4 & 42.0 & 41.0 & 37.9 & 30.2 & 32.0 & 42.5 & 40.9 & 41.0 & 38.6 & 39.3 & - \\ 
\system && A && 67.8 & 65.0 & 65.4 & 62.8 & 65.4 & 64.8 & 66.8 & 55.5 & 58.5 &    {63.6} & {3.4} & 
& {53.9} & {50.7} & {52.3} & {50.2} & 40.4 & 42.4 & {50.6}  & {51.5} & {52.1} & 44.5 & {48.9} & {9.6} \\

\cmidrule{1-1} \cmidrule{3-3} \cmidrule{5-15} \cmidrule{17-28}
Source  && B && 68.9 & 53.4	& 50.5	& 39.2 & 59.2 &	59.3 & 62.5 & 39.5 & 32.6 &	51.7 & - &
& 57.7 & 24.0 &	25.9 & 20.8 & 13.9 &  15.1 & 39.1	& 34.5	& 36.2 & 21.6 &	28.9 & - \\
\system &&  B && 69.6 & 65.3	& 66.2	& 63.9 & 66.5 &	66.7 & 69.2 & 53.7 & 59.0 &	64.5 & 12.8 &
&55.8 &51.6	&51.7	&49.8	&43.5	&48.6	&52.7	&51.1	&52.8	&46.0 & 50.4 & 21.5 
\\

\cmidrule{1-1} \cmidrule{3-3} \cmidrule{5-15} \cmidrule{17-28}

Source  && C && 68.3 &	66.1 &	66.0 &	58.2 &	66.4 &	65.8 &	68.3 & 53.4 & 53.2 &  62.8 & - &
& 57.3	& 50.6	& 51.4	& 47.2	& 36.4	& 39.0	& 53.2 & 52.2	& 53.1	& 43.6 & 48.4 &  - 
\\
\system &&  C &&70.7 & 69.5 &	69.8 & 67.9	 & 69.1 & 68.5 & 70.9 & 59.4 &	63.7 &	67.7  &  4.93 &
& 58.5	 & 56.1 & 55.9 & 54.2 & 42.6 & 46.1 &55.6 & 56.4 &56.6  & 50.8 & 53.3 & 4.9  \\

\bottomrule
\end{tabular}
}
\vspace{-0.1in}
\caption{\textbf{Results of different backbone networks used for adapting to changing environments.} Here, ``Source'' denotes directly applying the segmentation model to target tasks without adaptation.  A, B, C represent FCN-8s-ResNet101, DeepLab V3{$+$}~\cite{chen2018encoder}, and ResNet50-PSPNet~\cite{zhao2017pyramid}, respectively.}
\label{tbl:synthia}
\end{table*}

\subsection{Experimental Setup}
\noindent\textbf{Datasets and evaluation metrics}. Since our approach is designed to process different input distributions sharing the same label space for segmentation tasks, we use data with various weather and lighting conditions from \synthia~\cite{DBLP:journals/corr/RosSAW16},  a large-scale synthetic dataset generated with rendering engines for semantic segmentation of urban scenes.  We use {\scshape Synthia-Seqs}, a subset of \synthia showing the viewpoint of a virtual car captured across different seasons.  This dataset can be broken down into various weather and illumination conditions including ``summer'', ``winter'', ``rain'', ``winter-night'', \etc (See Table~\ref{tbl:synthia}). We consider two places from {\scshape Synthia-Seqs} for evaluation, \highway and \nyc, which contain 9 and 10 video sequences with different lighting conditions, respectively. We treat each sequence as a task, with around $1,000$ images on average, and each task is further split evenly into a training set and a validation set.

We first train a segmentation model using labeled images in the ``dawn'' scenario, and then adapt the learned model to the remaining tasks in each of the sequences in an unsupervised setting. During the adaptation process, following~\cite{DBLP:conf/iccv/ZhangDG17,hoffman2016fcns}, we only access labeled images from the first task (\ie, ``dawn''), and unlabeled images from the current task. To evaluate the performance of the segmentation model, we report mean intersection-over-union (mIoU) on the validation set of each task as well as the mean mIoU across all tasks.

\vspace{0.03in}
\noindent\textbf{Network architectures}. We use a pretrained VGG19 network as the encoder, and the architecture of the decoder is detailed in the supplemental material. We evaluate the performance of our framework with three different segmentation architectures, FCN-8s-ResNet101, DeepLab V3{$+$}~\cite{chen2018encoder}, and ResNet50-PSPNet~\cite{zhao2017pyramid}, which have demonstrated great success on standard benchmarks. FCN-8s-ResNet101 is an extension of FCN-8s-VGG network~\cite{long2015fully} that uses ResNet101 with dilations as the backbone, rather than VGG19. ResNet50-PSPNet contains a pyramid pooling module to derive representations at different levels that encompass sufficient context information~\cite{zhao2017pyramid}.  DeepLab V3{$+$}~\cite{chen2018encoder} introduces a decoder to refine the segmentation results along object boundaries.

\vspace{0.03in}
\noindent\textbf{Implementation details}. We use PyTorch for implementation and use SGD as the optimizer with a weight decay of $5\times 10^{-5}$ and a momentum of $0.99$. We set the learning rate to $10^{-3}$ and optimize for $10000$ iterations using standard SGD for training both source and target tasks. For fast adaptation with meta-gradients, we perform $50$ steps of meta updates.  We sample three source images in a mini-batch for training, and for each of these images from the source task we randomly sample two reference images, one from the current target task and one from the memory, as style references for generating new images. For style replay, the memory caches $100$ feature vectors per task representing style information from 100 target images.

\subsection{Results and Discussion}
\begin{figure*}[t!]
\begin{center}
   \includegraphics[width=0.9\linewidth]{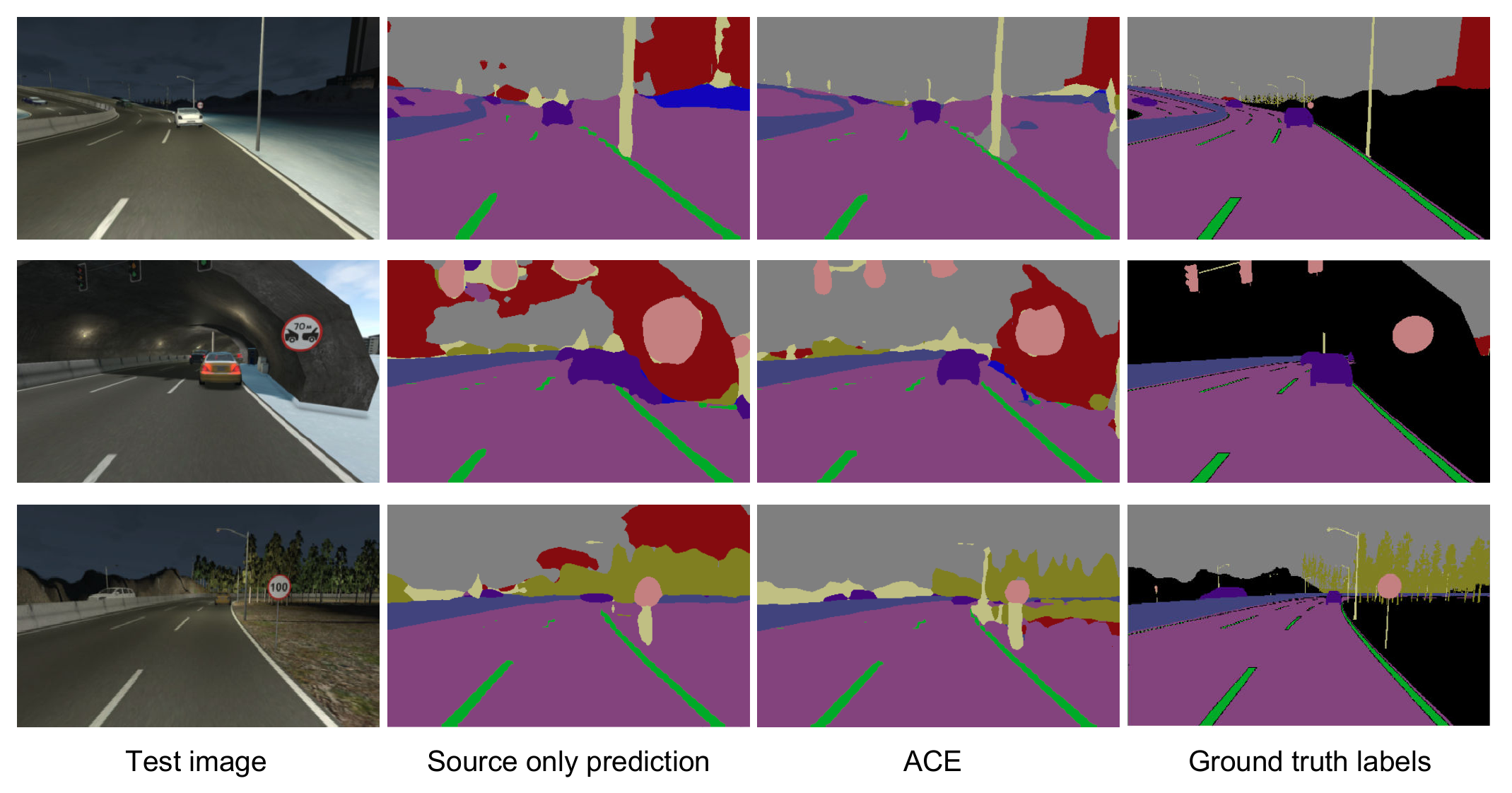}
\end{center}
\vspace{-0.3in}
\caption{\textbf{Visualizations of prediction maps by different methods}. Prediction results from sampled images using ResNet50-PSPNet and the corresponding source only model. The color black indicates a class to be ignored during training.}
\label{fig:prediction}
\end{figure*}

\noindent\textbf{Effectiveness of adapting to new tasks}. Table~\ref{tbl:synthia} presents the results of~\system and comparisons with source only methods, which directly apply the model trained on the source task to target tasks without any adaptation. We can observe that the performance of the source model degrades drastically when the distributions of the target task are significantly different from the source task (\ie, $15.4\%$ drop from ``dawn'' to ``'winter'' and $10.9\%$ drop from ``dawn'' to ``winter night'' with FCN-8s-ResNet101). On the other hand, \system can effectively align feature distributions between the source task and target tasks, outperforming source only methods by clear margins. For example, \system achieves a $3.4$ and $9.6$ (absolute percentage points) gain with FCN-8s-ResNet101 on \highway and \nyc, respectively. In addition, we can see similar trends using both ResNet50-PSPNet and DeepLab V3{$+$}, confirming the fact that the framework is applicable to different top-performing networks for segmentation. Comparing across different networks, ResNet50-PSPNet offers the best mean mIoUs on both datasets after adaptation. Although DeepLab V3{$+$} achieves the best results on the source task, its generalization ability is limited with more than $36.3\%$ performance drop when applying to the ``winter night'' task. However, \system can successfully bring back the performance with adaptation.
Furthermore, we also observe that the performance on \highway is higher than that on \nyc using different networks, which results from the fact the scenes are more cluttered with small objects like ``traffic signs'' in a city in contrast to highways. Figure~\ref{fig:prediction} further visualizes the prediction maps generated by \system and source only methods using ResNet50-PSPNet on \highway.
\begin{table}[h!]
\centering
\small
\ra{1}
\resizebox{1\linewidth}{!}{
\begin{tabular}{@{}*{9}c@{}}
\toprule
 &Method	&& Styles per task  &&\highway && \nyc \\
\cmidrule{2-2} \cmidrule{4-4} \cmidrule{6-6} \cmidrule{8-8} 
&	{Source} && -- && 60.2 && 39.3   	 \\
&\system && 0 && 61.2  && 46.2 \\ 
&\system && 200 &&  64.0  && 49.2 \\ 
\cmidrule{2-2} \cmidrule{4-4} \cmidrule{6-6} \cmidrule{8-8} 

&	    \system && 100 && 63.6  && 48.9 \\ 

\bottomrule
\end{tabular}
}
\vspace{-0.1in}
\caption{\textbf{The performance of \system depends on the number of exemplar style features stored in the memory unit.} The default number of features per task is 100, although we find that slight improvements can be made by increasing the number of stored vectors.  }
\label{tbl:memory}
\end{table}

\vspace{0.03in}
\noindent\textbf{Effectiveness of style replay}. We now investigate the performance of style replay using different numbers of feature vectors per task in the memory. Table~\ref{tbl:memory} presents the results. The accuracy of \system degrades by 2.4\% and 2.9\% on \highway and \nyc respectively when no samples are used for replay,  which confirms that style replay can indeed help revisiting previously learned knowledge to prevent forgetting. \system without reply is still better than source only methods due to the fact the segmentation network is still being updated with inputs in different styles. When storing more exemplar feature vectors (\ie, 200 per task) into the memory, \system can be slightly improved by $0.4\%$ and $0.3\%$ on \highway and \nyc, respectively. Here we simply use a random sampling approach to regenerate images in any of the historical styles, and we believe the sampling approach could be further improved with more advanced strategies~\cite{Castro_2018_ECCV}.

\vspace{0.03in}
\noindent\textbf{Comparisons with prior art}.
We now compare with several recently proposed approaches based on FCN-8s-ResNet101: (1)  {\scshape Source-Reverse} transfers testing images to the style of source images and then directly applies the segmentation model; (2) IADA aligns the feature distributions of the current task to those of a source task~\cite{wulfmeier2018incremental} in a sequential manner using adversarial loss functions~\cite{DBLP:conf/cvpr/TzengHSD17} such that the feature distributions can no longer be differentiated by a trained critic; (3) {\scshape ADDA-Replay} stores previous samples and prediction scores and uses a matching loss to constrain the segmentation outputs from previous tasks to remain constant as adaptation progresses. The results are summarized in Table~\ref{tbl:sota}. We can see that \system achieves the best results, outperforming other methods by clear margins, particularly on \nyc where a $6.9\%$ gain is achieved.

\begin{table}[h!]
\centering
\ra{0.9}
\resizebox{1.0\linewidth}{!}{
\begin{tabular}{@{}*{6}c@{}}
\toprule
 Method	&& \highway && \nyc \\
	\cmidrule{1-1} \cmidrule{3-3} \cmidrule{5-5} 
	{Source} && 60.2  && 39.3 	 \\
	{\scshape Source-Reverse} && 59.4  && 33.1 	 \\
	{\scshape IADA}~\cite{wulfmeier2018incremental} &&  60.7 &&  40.4 \\
	{\scshape ADDA-Replay}~\cite{bobu2018adapting} &&  61.9 && 42.0  \\
	\cmidrule{1-1} \cmidrule{3-3} \cmidrule{5-5} 
    \system && 63.6  && 48.9 \\ 
     \system-ADDA && 64.3  && 49.4 \\ 
\bottomrule
\end{tabular}}
\vspace{-0.1in}
\caption{\textbf{Comparisons with prior art}. We compare to two baseline methods, in addition to {\scshape IADA}~\cite{wulfmeier2018incremental} and {\scshape ADDA-Replay}~\cite{bobu2018adapting}.}
\label{tbl:sota}
\end{table}

Although {\scshape Source-Reverse} is a straightforward way to align feature distributions, the performance is worse than directly applying the source model. We suspect that this performance drop occurs because of small but systematic differences between the original source data on which the segmentation engine was trained, and the style transferred data on which no training ever occurs.
In contrast, \system trains the segmentation network on synthesized images, and constrains the segmentation output on generated images to be compatible with output on the original source image. In addition, {\scshape IADA} improves the source only model slightly by aligning feature distributions in a sequential manner, however, it relies on an adversarial loss function that is hard to optimize~\cite{arjovsky2017towards}. More importantly, while {\scshape IADA} proves to be successful for classification tasks, for tasks like segmentation where multiple classifiers are used for deep supervision~\cite{zhao2017pyramid,long2015fully} at different distance scales, it is hard to know which feature maps to align to achieve the best performance. Further, we can also see that {\scshape ADDA-Replay} offers better results compared to {\scshape IADA} by using a memory to replay, however this requires storing all samples from previous tasks.  

\begin{figure}[t!]
\begin{center}
   \includegraphics[width=1.0\linewidth]{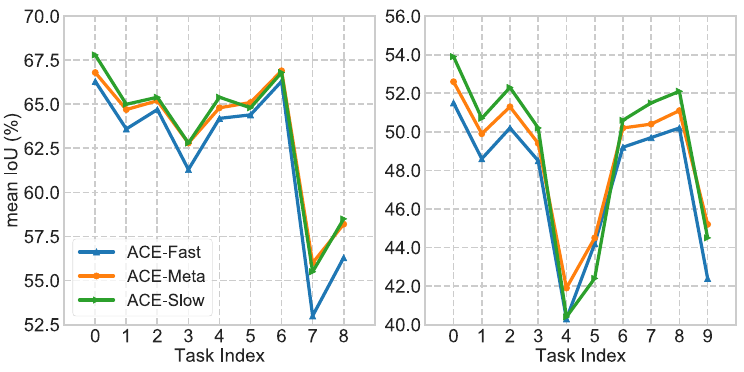}
\end{center}
\vspace{-0.2in}
\caption{\textbf{Results of fast adaptations using \system-Meta}. \system-Slow: full batch training with SGD for $10$K iterations. \system-Fast: batch training with SGD using 600 iterations (the same number as \system-Meta).}
\label{fig:adaptation}
\end{figure}

Note that ADDA~\cite{DBLP:conf/cvpr/TzengHSD17} focuses on aligning distributions at the feature-level rather than the pixel-level, and this reduces low-level discrepancies in our approach. Yet, our approach is complimentary to approaches that explore feature-level alignment in the segmentation network at the cost of storing image samples for replay. When combining ADDA with \system, $0.7\%$ and $0.6\%$ further improvements are achieved on both \highway and \nyc.

\vspace{0.03in}
\noindent\textbf{Fast adaptation with meta-updates}. \system achieves good results by batch training on each task using tens of thousands of SGD updates. We are also interested in how to adapt to the target task quickly by leveraging recent advances of meta-learning. We propose the meta-update method (\system-Meta) which uses Reptile for learning meta-parameters, which are then fine-tuned to a specific task using only $600$ iterations of SGD.  We compare to \system-Fast, which also uses $600$ iterations per task, but without meta-learning, and also \system-Slow, which uses full batch training with SGD for $10$K iterations. The results are summarized in Figure~\ref{fig:adaptation}. \system-Meta achieves better performance compared to \system-Fast, trained under the same settings, almost for all the target tasks on both \highway and \nyc, and we observe clear gains when applying the model to ``winter'' and ``winter night''. Moreover, the results of \system-Meta are on par with full batch training with SGD, demonstrating that meta-updates are able to learn the structures among different tasks.

\begin{figure}[t!]
\begin{center}
   \includegraphics[width=0.83\linewidth]{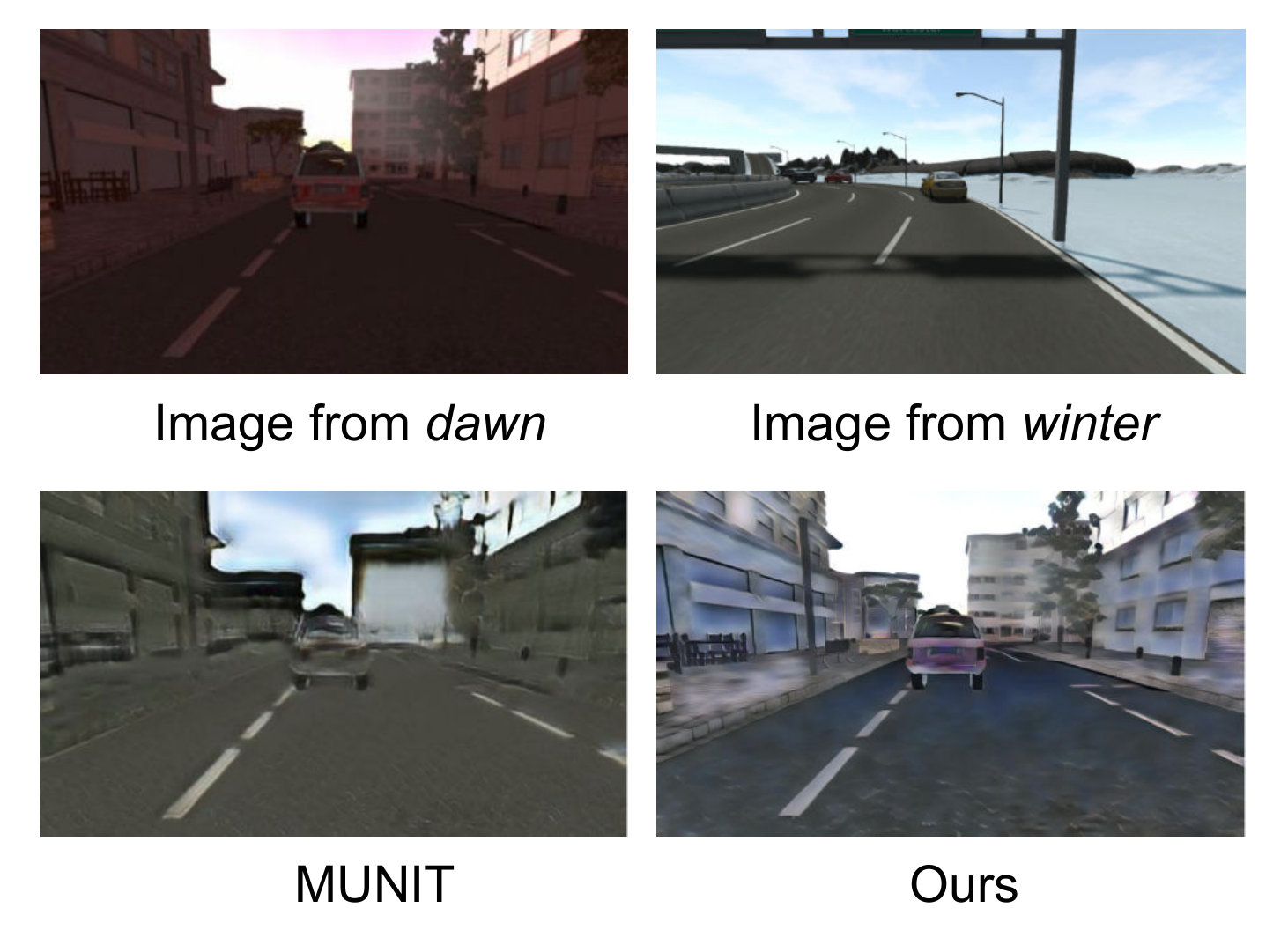}
\end{center}
\vspace{-0.3in}
\caption{\textbf{Comparisons with different image synthesis methods}. Images generated with our method and comparisons with MUNIT by transferring the image from ``dawn'' to ``winter''.}
\label{fig:gans}
\end{figure}

\vspace{0.03in}
\noindent\textbf{Image generation with GANs}. We compare images generated by \system to MUNIT~\cite{huang2018multimodal} in Figure~\ref{fig:gans}. MUNIT learns to transfer the style of images from one domain to another by learning a shared space regularized by cycle consistency, and compared to CycleGAN~\cite{DBLP:conf/iccv/ZhuPIE17}, it is able synthesize a diverse set of results with a style encoder and a content encoder that disentangle the generation of style and content. Note that MUNIT also relies on AdaIN to control the style, but uses a GAN loss for generation. We can see that image generated with our approach preserves more detailed content (\eg, facade of the building), and successfully transfers the snow to the walkway, while there are artifacts (\eg, blurred regions) in the image generated with MUNIT. 

\section{Conclusion}
We presented \system, a framework that dynamically adapts a pre-trained model to a sequential stream of unlabeled tasks that suffer from domain shift. \system leverages style replay to generalize well on new tasks without forgetting knowledge acquired in the past. In particular, given a new task, we introduced an image generator to align distributions at the pixel-level by synthesizing new images with the contents of the source task but in the style of the target task such that label maps from source images can be directly used for training the segmentation network. These generated images are used to optimize the segmentation network to adapt to new target distributions. To prevent forgetting, we also introduce a memory unit that stores the image statistics needed to produce different image styles, and replays these styles over time to prevent forgetting. We also study how meta-learning strategies can be used to accelerate the speed of adaptation. Extensive experiments are conducted on \synthia and demonstrate that the proposed framework can effectively adapt to a sequence of tasks with shifting weather and lighting conditions. Future directions for research include how to handle distribution changes that involve significant geometry mismatch.

{\small
\bibliographystyle{ieee}
\bibliography{reference}
}

\end{document}